\documentclass[letterpaper, 10 pt, conference]{ieeeconf}  

\IEEEoverridecommandlockouts                              

\usepackage{adjustbox}
\usepackage{graphicx}
\usepackage{mdframed}
\usepackage{textcomp,amsmath,longtable}
\usepackage{latexsym}
\usepackage{placeins}
\usepackage{float}
\usepackage{ltablex,array}
\usepackage{multirow}

\usepackage[figuresright]{rotating}

\usepackage{tabularx}
\usepackage{makecell}
\title{\LARGE \bf Social Robot Scenarios for Real-World Child and Family Care Settings through Participatory Design}

\author{Anouk Neerincx$^{1}$
\thanks{$^{1}$ Department of Information and Computing Sciences, Utrecht University \tt\small{a.neerincx@uu.nl}}%
}

\begin{document}

\maketitle
\thispagestyle{empty}
\pagestyle{empty}

\begin{abstract}

This paper discusses a 5-year PhD project, focused upon the implementation of social robots for general child and family care settings in the Netherlands. The project is a collaboration with general Dutch family care organisations as well as specialized child mental health care organisations. The project adapts a bottom-up, participatory design approach, where end users are included in all stages of the project. End users consist of children, parents, and family care professionals, who all have different needs, regarding the social robot behaviors as well as the participatory design methods. This paper provides suggestions to deal with these differences in designing social robots for child mental support in real-world settings.

\end{abstract}


\section{PROBLEM STATEMENT AND COMMUNITY}

Social robots show potential in the field of child education \cite{belpaeme2018social} and child health care \cite{dawe2019can}. Previous research shows that children generally like to work with robots. Robots can increase engagement in children during treatment, reduce stress, and robots can be used as a multimedia tool. Social robots can therefore make healthcare treatments as well as education more enjoyable for children. Additionally, a social robot can help with establishing a connection between the child and the therapist and/or teacher, for example by creating a safe environment for the child, stimulating self-disclosure. 
Mental as well as general health care organizations in the Netherlands express the need to make use of the potential of technological solutions that exist (such as social robots) for child mental support in various care and educational settings \cite{neerincx2021social}. Before implementing such solutions, it is important to first design those solutions user-centered, before studying the effects and risks thoroughly (long-term) in real-world settings. The discussed 5-year PhD project (started November 2019), focuses on designing and testing social robot behaviors for child mental support, together with several Dutch health care organizations, primary schools, and science festivals

\subsection{Organisations}
Several organisations are involved in this PhD project. First of all, the Dutch Child and Family Center (CJG), which focuses on providing general health care (e.g., vaccinations, eye tests) as well as family coaching. Second, Levvel, which provides specialized mental health care, such as improving social-emotional skills. Third, several primary schools are involved by organizing robot co-design workshops for their students. Finally, we collaborate with several science festivals, to co-design and test social robots in public spaces.

\subsection{Research Questions}
The two main research questions studied in this PhD project are:

\begin{itemize}
    \item How can a social robot evoke appropriate self-disclosures (e.g., to facilitate the connection between child and professional, to support social-emotional skills)?
    \item How can a social robot reduce stress during treatments?
\end{itemize}

The main target group of our research is children aged 8 to 12 years old, since this age group is suitable for working with robots, and it is the biggest client group of our collaborators.

\section{METHODS \& PRELIMINARY RESULTS}

\subsection{Dutch Child and Family Center}

\subsubsection{Exploration}
In an exploration phase, the Dutch Child and Family Center included perspectives of child professionals (N=12) and parents (N=5), to explore different scenarios and user requirements. Four focus groups were carried out of 1,5 hours each. The focus group methodology was based on grounded theory, and included open-ended questions as well as card sorting exercises.

Main examples of user requirements are:

\begin{itemize}
    \item Robot shall adapt its behavior to the child’s needs
    \item Robot shall be safe to use (data storage, system quality)
    \item Robot shall enrich the interaction by means of positive reflections (e.g., feedback), “small talk”, and games
\end{itemize}

Main examples of scenarios are:

\begin{itemize}
    \item Robot as icebreaker
    \item Robot to assist in communicating
    \item Robot to make the treatment more fun
\end{itemize}

We also carried out a child robot co-design workshop trial at the Dutch Child and Family Center (N=4), including creative activities such as theatre play, drawing, and playing with different robots. After this trial, several workshops were organized at different primary schools as well as at our university.

\subsubsection{Testing}
The designs found in the exploration phase are currently being tested and evaluated in (low-risk) real-world settings, for example at vaccination days. Standardized questionnaires, thematic analysis of open-ended questions, as well as video analysis are being used in analyzing the effects. First results of our studies show for example that robots can reduce stress by increasing engagement (but can hinder parent participation) \cite{neerincx2021waiting}, the children evaluate the use of social robots positively, girls trust the robot more than boys, and children can learn from the robot.

\subsection{Primary Schools}

Four different robot co-design workshops have been organized at primary schools, with children age 8-12 (N=50). The activities consisted of drawing or writing, theatre play, and robot programming. Results include user requirements and scenarios. 
The most popular scenarios were:

\begin{itemize}
    \item Social robot at vaccination day
    \item Social robot to assist at school / with homework
    \item Social robot at an eye test 
\end{itemize}

The workshop was evaluated by a homework assignment from the teacher, where she asked: What did you learn? What did you like? The children stated that they mostly learned about how robots work and about programming. They stated that they liked the theatre activity the most.

\subsubsection{Summer School}
Currently, an adapted version of the robot co-design workshop is being carried out at a summer school at Utrecht University (expected N=500). In this workshop, the children are asked to make a robot drawing in small groups based on one scenario (from a set of 7 based on our previous research), online programming of a robot, and meeting a robot in our lab. Features of the drawings as well as online programming activity will be analyzed based on a coding scheme.

\subsection{Levvel}

In the collaboration with Levvel, focus groups were done with several child mental health care professionals (N=8), to explore the potentials of social robot exercises to improve social-emotional skills. Based on standardized social-emotional exercises provided by the experts, social robot exercises are currently under development, which will be tested in real-world settings.

\subsection{Science Festivals}

Collaborations with science festivals study the topic of self-disclosure in a first real-world robot encounter. At Betweter science festival, we studied the effect of self-disclosure category, human personality and robot identity on adults (N=80) self-disclosure by means of standardized questionnaires and audio analysis (full paper at RO-MAN 2022). At expeditie NEXT science festival for children (N=83), we did a similar study, but we added a co-design activity, where children could fill out a comic strip of a child-robot interaction in either an education or health care setting. Results are currently being analyzed.

\section{DISCUSSION: WHY PARTICIPATORY DESIGN, AND EXPERIENCE}

Especially when designing technological solutions for in practice, it is important to incorporate the views and needs of the end users. Since the researchers and designers are no experts of the specific field in which the technology will be used, they will probably have unexpected views and needs, which need to be taken into account for a satisfactory product. Also, in the field of social robots, often new robots are being designed which are only used for a certain period of time (novelty effect), then cast aside, after which a new model is developed. Designing together with the end users might encourage more long-term, sustainable use. By using participatory design methods, you are making sure that the right product is being designed. It is important though to include all end users in the process, and to adapt the participatory design methods to the target group. For example, children can benefit from multiple ways of expressing themselves (besides verbally). Self-expession might also be culture-related \cite{neerincx2016child}. Therefore, including more creative methods such as drawing and theatre play might be beneficial. Creative methods might stimulate adults to think "out-of-the-box" as well. By first using participatory design, and then use more 'standardized' methods to test those designs, you use the best of both worlds.
The discussed PhD project was started at the end of 2019 as a bottom-up, participatory design project. I started my participatory design experience in the beginning of 2020 by organizing focus groups.

\bibliography{literature.bib}
\bibliographystyle{IEEEtran}

\end{document}